\documentclass{llncs}
\usepackage{makeidx}  
\usepackage{graphicx}
\usepackage{mathtools}
\usepackage{color}
\usepackage{amssymb}
\usepackage{amsmath}
\usepackage{graphicx}
\usepackage{epstopdf}
\usepackage{supertabular}
\usepackage{array}
\usepackage{multirow}
\usepackage{setspace}
\usepackage{booktabs}
\usepackage{subfigure}

\usepackage[normalem]{ulem}

\begin{document}
\mainmatter              
\title{Automated Segmentation of Knee MRI using Hierarchical Classifiers and Just Enough Interaction based Learning: Data from Osteoarthritis Initiative}
\titlerunning{Learning Based LOGISMOS Knee Segmentation}  
\authorrunning{Kashyap et. al}   

\author{Satyananda Kashyap \inst{1} \and Ipek Oguz\inst{1,2}  \and Honghai Zhang\inst{1}  \and Milan Sonka\inst{1}}
\institute{Iowa Institute for Biomedical Imaging, The University of Iowa, Iowa City IA, USA\\
\and 
Department of Radiology, University of Pennsylvania, Pennsylvania,  PA, USA \\
\email{satyananda-kashyap@uiowa.edu}}
\maketitle              

\begin{abstract}
We present a fully automated learning-based approach for segmenting knee cartilage in presence of osteoarthritis (OA). The algorithm employs a hierarchical set of two random forest classifiers. The first is a neighborhood approximation forest, the output probability map of which is utilized as a feature set for the second random forest (RF) classifier. The output probabilities of the hierarchical approach are used as cost functions in a Layered Optimal Graph Segmentation of Multiple Objects and Surfaces (LOGISMOS). In this work, we highlight a novel post-processing interaction called just-enough interaction (JEI) which enables quick and accurate generation of a large set of training examples. Disjoint sets of 15 and 13 subjects were used for training and tested on another disjoint set of 53 knee datasets. 
All images were acquired using double echo steady state (DESS) MRI sequence and are from the osteoarthritis initiative (OAI) database. Segmentation performance using the learning-based cost function showed significant reduction in segmentation errors ($p< 0.05$) in comparison with conventional gradient-based cost functions.

\end{abstract}

\section{Introduction}
Osteoarthritis (OA) is one of the most prevalent aging diseases in our society \cite{jama1990}. Early detection of structural differences associated with OA is important for testing interventional drugs in clinical trials. Segmentation is a crucial first step in structural analysis. Manual segmentation takes several hours of effort and is prone to inter/intra-observer variability and operator-induced bias. Reproducible automated segmentation is challenging in presence of severe OA due to cartilage thinning, osteophytes, bone marrow, and cartilage lesions in the MR volume. 

Several automated algorithms for knee segmentation have been proposed in the literature such as approximate binary k-NN classifiers \cite{Folkesson2007}, deformable active shape models \cite{Fripp2010}, local image patch optimization using Markov random fields \cite{Lee2011}, hierarchical two stage cartilage classifiers optimized by graph cuts \cite{Wang2014}, and multi-atlas labeling with locally weighted voting \cite{Lee2014}. Many of these methods make use of locally (not globally) optimal techniques to solve the segmentation problem. For example, Wang et al.\ \cite{Wang2014} used multi-label graph cuts to optimize the background combining bone regions with true background and the cartilage classifier outputs. LOGISMOS algorithm \cite{Yin2010} can simultaneously segment multiple surfaces in different objects taking contextual information between them to make the segmentation robust with guaranteed global optimality.

Severe OA pathology alters the tissue appearance substantially, limiting the segmentation performance when using simple 
cost functions. A major limiting factor of learning-based costs (used in \cite{Kashyap2013,Yin2010,Folkesson2007,Wang2014}) is the time consuming task of curating a large numbers of accurate training examples. Several interactive correction methods were designed to ease the post-processing corrections such as thin plate splines \cite{ross2009lung}, live-wires \cite{schenk2000efficient} and active shape model based interactions \cite{schwarz2008interactive}. In these cases and others, the refinement mainly corrects the surface errors directly to match the object boundaries and/or are prohibitively expensive to be computed in real time.  

We present a fully automated LOGISMOS segmentation algorithm based on learned costs using a two-stage hierarchical random forest classifiers (RF). Unlike in \cite{Wang2014}, our method uses two variations of RF classifier, the first being a neighborhood approximation forests (NAF) \cite{konukoglu2013} followed by an RF classifier on a geometric graph thereby learning from a combination of local and global context and textural features. JEI approach \cite{sun2013} was used to prepare the training data, substantially reducing the interaction time compared to traditional voxel-by-voxel post-processing approaches. This is achieved by interacting with the underlying graph algorithm. While similar to \cite{boykov2001interactive}, their graph-cut interaction approach for multiple objects does not guarantee global optimality. Live-wires with embedded interaction capabilities have a similar drawback of being unable to maintain global optimality for multiple surfaces and objects. LOGISMOS-JEI handles multiple object interactions while maintaining global optimality.

\section{Methods}
\begin{figure}[htb]
	\centering
	\includegraphics[width=0.7\textwidth]{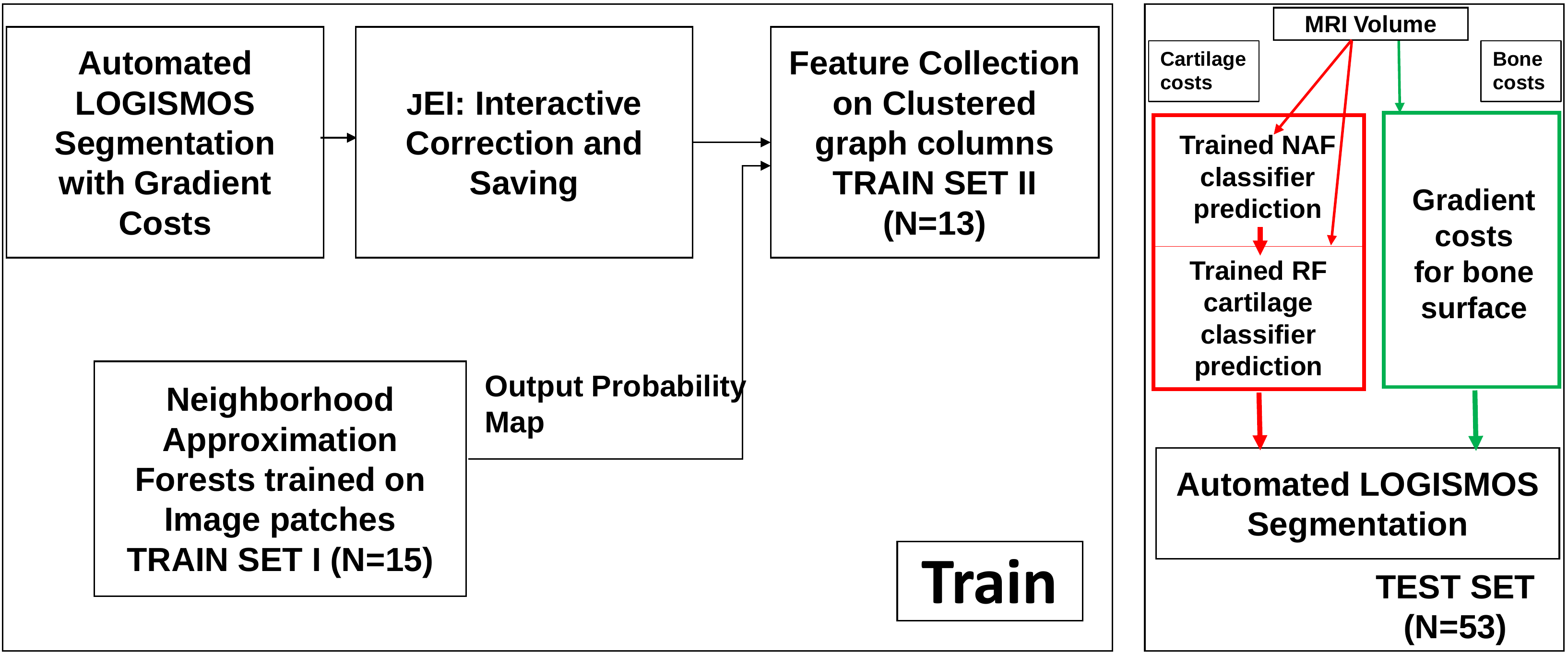}
	\caption{The learning-based segmentation algorithm workflow.}
	\label{fig:workflow}
\end{figure}

The proposed segmentation work-flow is outlined in Fig.\ \ref{fig:workflow} beginning with an automated LOGISMOS segmentation using gradient-based costs. The optimization finds the minimum closed set on a node weighted graph. The optimized residual graph and the image volume are loaded into the custom built graphical user interface to examine the segmentation quality and perform JEI. Upon completing JEI, the final edited surfaces are used as training examples for the hierarchical classifier system. Upon training, the classifier probability values on a test volume is assigned as cartilage node costs in LOGISMOS graph. For bone surface segmentation, the initially-employed gradient-based costs were very robust and remained unchanged.
\subsection{LOGISMOS Segmentation Algorithm}

The algorithm segments multiple objects and surfaces simultaneously in a graph based framework. Initialization of the algorithm consists of volume of interest (VOI) detection using an AdaBoost classifier \cite{Freund1997} trained on manually identified VOI's. Training used 9 different 3D Haar-like features applied at different scales. The smaller region localized by the VOI helps reduce the computation time. Further, these VOI bounds are used for fitting the mean shape mesh $S_0$ for each bone (femur and tibia respectively) using affine registration. A patient-specific bone shape $S$ is important since the final segmentation of both the bone and the cartilage surfaces is defined by this shape prior. $S$ is obtained by a single-surface single-object LOGISMOS segmentation using $S_0$.

A geometric graph is constructed representing each surface of the object to be segmented. Non-intersecting graph column construction is a crucial step in ensuring topologically correct segmentation. This is enforced by mimicking the behavior of electrically charged particles resulting in an electric lines of force (ELF) based graph column construction. All graph nodes are assigned unlikeliness costs based on the local image intensity gradient. To represent the segmentation task as a max-flow problem, columns are connected by intra- and inter-column arcs enforcing surface smoothness constraints. The final multi-object multi-surface segmentation add additional arcs to enforce inter-object and inter-surface constraints whose arc construction and final solution are described in \cite{Yin2010}. A 1D derivative operator, $\nabla(x,y,z)$, along the column direction was used to determine bone-surface costs while the cartilage costs employed  weighted (determined empirically) first and second order derivatives: $ w * \nabla(x,y,z) + (1-w)* \nabla^2(x,y,z)$.

\subsection{Just Enough Interaction}
Following LOGISMOS segmentation, JEI editing on the resulting surfaces is done as needed. The image volume, residual graph and the ELF based geometric graph are loaded into a custom designed GUI. The previously reported JEI methods \cite{sun2013} were extended from a single object, two surface interaction to a multi-object multi-surface interaction for knee MRI. A $k$-dimensional tree based interaction algorithm was designed for cost modification based on user inputs. Furthermore, a faster max-flow optimization algorithm was utilized for immediate feedback on the interaction. The JEI workflow was as follows: 1) specifying approximately correct boundary points, referred to as \emph{nudge points} hereafter,  on a chosen 2D slice, 2) modification of local graph costs in the entire 3D neighborhood of interaction, 3) max-flow re-computation in 3D, resulting in corrected surfaces within milliseconds. The process was repeated until the results were deemed satisfactory.

\subsubsection{Underlying Interaction and Graph Cost Modification}
The user specified nudge points form a 3D contour and its intersection (intersecting nodes) with the graph columns is identified by utilizing a $k$-dimensional tree that stores the coordinates of all graph nodes to enable $O(\log n)$ query on $N$ nearest (determined empirically) nodes for any given node. The node costs (i.e. unlikeness) of all columns with intersecting nodes are modified as:
\[
c(i,j)= 
\begin{dcases}
0, & \text{if } D((i,j),n(i,j)) < \Delta\\
1, & \text{otherwise}
\end{dcases} \; ,
\]
where $c(i,j)$ is the cost of node $j$ on column $i$, $D((i,j),n(i,j))$ is the distance between node $(i,j)$ and its nearest intersecting node $n(i,j)$, and $\Delta$ is a tolerance. The max-flow optimization then continues from its previous state using the updated costs and consequently produces the updated surfaces.

\subsection{Classifier System Design}
Two RF based classifiers in hierarchy were used to train cartilage regions. A NAF \cite{konukoglu2013} trained on example image patches was used as the first stage followed by a second RF classifier \cite{breiman2001random} with features collected along the ELF based geometric graph nodes. The output probability maps of the NAF were not directly input as costs to the graph optimization. They were used with other image based features for training of the second RF classifier. The advantage of this approach is that information is gathered from a larger global neighborhood in combination with local features. The NAF classifier gathers contextual and textual information from a larger neighborhood of 3D image patches. The RF classifier collects local feature information along the geometric search columns of the graph. Disjoint training sets were used to help build a more realistic RF model based on actual NAF performance on unseen images.

\subsubsection{Neighborhood Approximation Forests}
\begin{figure}[htb]
	\centering
	\includegraphics[height=2.3cm]{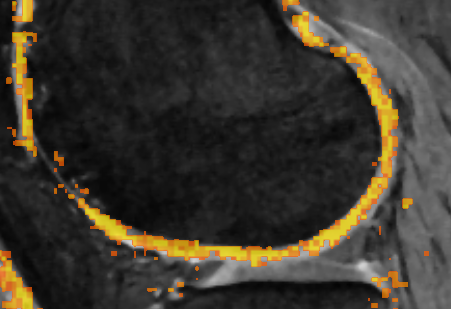}
	\includegraphics[height=2.3cm, width=3.3cm]{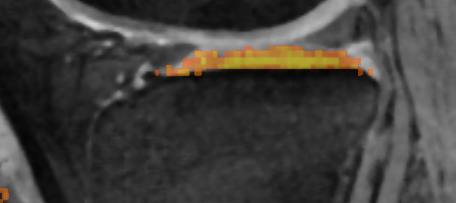}
	\caption{The output probability map of the NAF for an unseen image overlaid on the image volume. The color map indicates the probability output values with brighter color indicating higher probability of the voxel being a cartilage region.}
	\label{fig:NAF}
\end{figure}

NAF is based on a random forests framework which approximates the nearest neighbors of image patches based on a user defined distance function to optimize the node split. The pairwise distance function $\rho(I,J)$ between each of the training image patches is defined as $\rho(I,J) = \parallel seg(I) - seg(J) \parallel_{l_0}$ where $seg(.)$ is the segmentation label map for the corresponding image patch. Intuitively it measures similarity between image patches based on the segmentation. The algorithm learns to group image patches which appear similar to each other based on this neighborhood distance criteria. The output probability map of the unseen image (Fig.\ \ref{fig:NAF}) was used as one of the inputs for a second RF classifier. 

\subsubsection{Clustered Random Forest Classifier}
The second RF was trained on features collected at each node of the geometric graph. JEI edited bone mesh surfaces were used for geometric graph construction during training. Positive example labels corresponded to the nearest cartilage mesh intersection along each search column. The different features collected at each node point are shown in Table \ref{featureLabel} with feature values interpolated to the search path points from corresponding feature volumes. To handle the large variability of cartilage intensities in the volume, a k-means clustering algorithm was applied to the $S_0$ mesh of femur and tibia respectively resulting in spatial parcellation of the pre-segmented mesh surfaces into 40 clusters each (total 80). The clustering trained regionally-specific appearance models by accounting for the surrounding menisci, muscle, bone and other anatomies. The probability response to the features along the search nodes in the testing datasets provided the node costs for graph optimization instead of gradient based costs. 

\begin{table}[htb]
	\centering
	\caption{A list of features used to train the second RF classifier. }
	\label{featureLabel}
	\begin{small}
		\begin{tabular}{l|l}
			\hline
			{\bf Index} & {\bf Description} \\ \hline
			1--9           & 3 eigenvalues of Hessian matrices on intensity image at $\sigma=0.5,1.0,2.0$ mm\\
			10--15         & 1st Gaussian gradient on intensity and NAF probability volumes \\
			& at $\sigma=0.36,0.7,1.4$ mm\\
			16--18         & Intensity, Gaussian smoothed intensity, and NAF probability volumes \\
			19--20         & Laplacian derivative of intensity volume at $\sigma=0.36,0.7$ mm \\
			21             & Gabor texture feature \\
			22--25         & Intensity statistics: mean, variance, skewness and kurtosis of a \\
			& 2 $mm^3$ region centered around each graph node\\
			26--28         & Haar features (1.5mm kernel) along horizontal, vertical \& diagonal directions \\
			29--30         & 1D directional gradient along the search column direction on NAF \\
			& probability and intensity volume\\
			\hline
		\end{tabular}
	\end{small}
\end{table}

\section{Experimental Methods}
MRI volumes with independent standards are available from the OAI. All subjects were imaged using a DESS protocol with a voxel resolution of $0.36\times0.36\times0.7$ mm$^3$. All MR volumes in this study were from diseased subjects. The data were divided into two training sets with 15 and 13 which were used to train the NAF and the second RF classifier. The data-sets used for training the clustered RF classifier were first inspected and JEI edited. 53 data-sets were used for testing.

All image volumes were first LOGISMOS segmented using gradient costs. The geometric graph had 8006 and 8002 graph columns for the femur and tibia objects respectively. The graph parameters are listed in Table \ref{graph-parameters}.

\begin{table}[htb]
	\centering
	\caption{Parameters used for graph construction. Minimum inter-surface and inter-object separations are zero.}
	\label{graph-parameters}
	\begin{tabular}{lccccc}
		\hline
		& Inter-surface & Inter-object & Smoothness & Column size & Node spacing \\
		& max (nodes)   & max (nodes)  & (nodes)    & (nodes)     & (mm) \\
		\hline
		Learned cost  & 40            & 120          & 4          & 121         & 0.15 \\
		Gradient cost & 20            & 60           & 2          & 61          & 0.20 \\
		\hline
	\end{tabular}
\end{table}

The NAF features consisted of image patches sampled over 15 data-sets with 1521 sample points per patch. Because of the highly imbalanced ratio between the negative and positive labels, we considered a neighborhood around the cartilage labels and marked them as negative examples. The image patches collected consisted of all the positive and the surrounding negative labels. We trained a set of 200 trees with 40,000 images patches as inputs to each tree.

The second set of RF classifier was trained on 13 JEI-corrected data-sets with 30 features (see Table \ref{featureLabel}) along with the ELF search path for each node. 80 ($40\times2$) RF classifiers were trained with each one representing the given cluster with 800 trees per forest.

\section{Results}

Surface positioning errors (compared against independent standard) achieved by hierarchical classifier, gradient cost and single stage RF classifier are listed in Table \ref{error}. It shows a significant reduction in signed errors for the femur using learned costs ($p<0.001$). While the signed tibia errors were not statistically significantly different ($p=0.62$, difference between means of 1/18 voxel), the error variance was substantially reduced showing that larger errors were avoided by the new strategy. The reduction in unsigned surface positioning errors was significant for both the femur and tibia ($p<0.001$). 
Table \ref{error} also shows the benefits of the hierarchical classifier when compared with the single stage RF classifier. Although the single stage RF improves the segmentation accuracy, the addition of the hierarchical NAF stage further improves the accuracy with the largest reduction in error seen in the tibial regions.

\begin{figure}[htb]
	\centering
	\subfigure[]{\includegraphics[height=4.0cm]{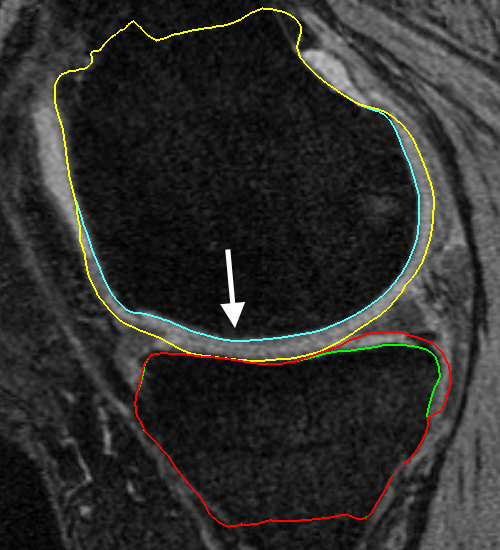}}
	\subfigure[]{\includegraphics[height=4.0cm]{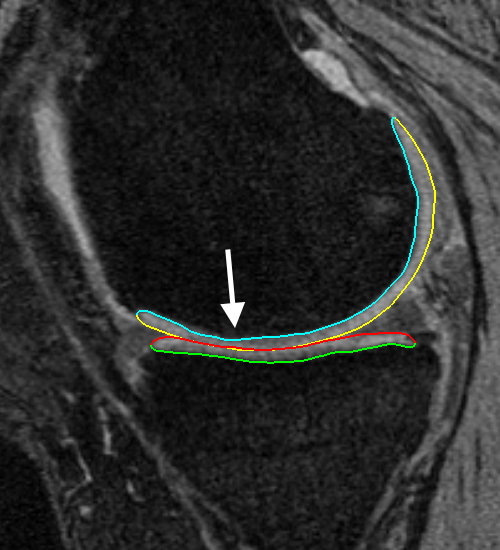}}
	\subfigure[]{\includegraphics[height=4.0cm]{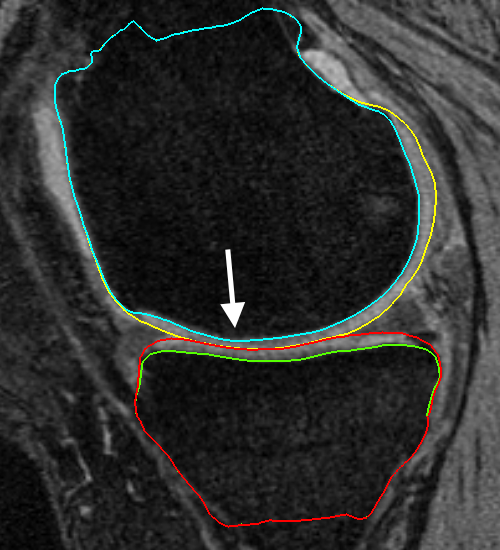}}
	\caption{Segmentation accuracy in a representative subject.
		(a)~Gradient-costs LOGISMOS segmentation.
		(b)~Independent standard.
		(c)~Learned-costs LOGISMOS segmentation. Regions marked by the arrow shows clear improvement in the segmentation quality when using the learned costs.}
	\label{fig:qual}
\end{figure}

Fig.\ \ref{fig:qual} qualitatively compares the segmentation accuracies between the two methods and the independent standard. Both the femur and tibia are shown with their respective bone and cartilage segmentations showing good agreement between learning-based segmentation and the independent standard. 

\begin{table}[htb]
	\centering
	\caption{Border positioning errors (mm) achieved by hierarchical classifier, gradient cost and single stage RF classifier.}
	\label{error}
	\begin{tabular}{lccc|c}
		\hline
		& NAF+RF (\bfseries{Proposed})            & Gradient          & $p$-value & RF only        \\ 
		\hline                                           
		Femur signed   & \bf 0.03$\pm$0.19 &-0.31$\pm$0.28     & $\ll0.001$  & -0.06$\pm$0.18 \\ 
		Femur unsigned & \bf 0.55$\pm$0.11 & 0.68$\pm$0.20     & $\ll0.001$  &  0.56$\pm$0.11 \\ 
		\hline                                           
		Tibia signed   &     0.10$\pm$0.17 & 0.08$\pm$0.35 & ~~~0.62      & 0.16$\pm$0.24  \\
		Tibia unsigned & \bf 0.61$\pm$0.14 &     0.81$\pm$0.18 & $\ll0.001$  & 0.65$\pm$0.17  \\
		\hline
	\end{tabular}
\end{table}

\section{Conclusion}
A novel fully automated learning-based algorithm for designing cost functions used in LOGISMOS segmentation was presented. Cost function was designed by optimizing a hierarchical set of classifiers, each associated with one of two different learning methods. We also demonstrated the use of JEI for acquiring training examples. The presented method was highly accurate when compared to pre-JEI results. Understanding the effect of feature selection and the mutual combination of parameters from both classifiers remains as future work. 
\section*{Acknowledgment}
OAI support for providing the MRI volumes and manual segmentation gratefully acknowledged. This research was supported by NIH grant - R01EB004640.  Computation support through Helium cluster provided by University of Iowa.

\bibliographystyle{splncs03}
\bibliography{paper637}
\end{document}